\documentclass[nohyperref]{article}
\pdfoutput=1

\usepackage{amsmath,amsfonts,bm}

\def\eqref#1{equation~\ref{#1}}

\def\1{\bm{1}}

\DeclareMathAlphabet{\mathsfit}{\encodingdefault}{\sfdefault}{m}{sl}
\SetMathAlphabet{\mathsfit}{bold}{\encodingdefault}{\sfdefault}{bx}{n}

\usepackage{microtype}
\usepackage{efbox,graphicx}
\usepackage{subfigure}
\usepackage{booktabs} 
\usepackage{adjustbox}

\usepackage{hyperref}

\usepackage[preprint]{icml2022}

\usepackage[capitalize,noabbrev]{cleveref}
\usepackage{amsmath,amsthm,amssymb}
\usepackage{mathtools}

\theoremstyle{plain}
\newtheorem{theorem}{Theorem}[section]
\newtheorem{proposition}[theorem]{Proposition}

\newtheorem{corollary}[theorem]{Corollary}
\theoremstyle{definition}

\theoremstyle{remark}

\usepackage{graphicx}
\usepackage{subfigure}  
\usepackage{todonotes}
\usepackage{dsfont}
\usepackage{tabulary}
\usepackage{tabularx}
\usepackage{makecell}
\usepackage{wrapfig}
\usepackage{fancyvrb}
\usepackage{MnSymbol}%
\usepackage{wasysym}%
\usepackage{multirow} 
\usepackage{xspace}
\newcommand{\ie}[0]{\emph{i.e.},~}
\newcommand{\eg}[0]{\emph{e.g.},~}
\newcommand{\aka}[0]{a.k.a.~}

\newcommand{\wrt}{w.r.t.~}

\newcommand{\defeq}{\ensuremath{\doteq}}
\usepackage{dsfont}

\newcommand{\set}[1]{\ensuremath{{#1}}}
\newcommand{\gr}[1]{\ensuremath{{#1}}}

\newcommand{\tuple}[1]{\ensuremath{({#1} )}}
\newcommand{\Reals}{\mathds{R}}
\newcommand{\Complex}{\mathds{C}}

\newcommand\norm[1]{\left\lVert#1\right\rVert}
\newcommand\btuple[1]{\big(#1\big)}

\newcommand{\Gg}{\gr{G}\xspace}
\renewcommand{\gg}{{g}}
\newcommand{\egr}{e}

\newcommand{\Xs}{\set{X}}
\newcommand{\Zs}{\set{Z}}
\newcommand{\Ys}{\set{Y}}

\newcommand{\Ds}{\set{D}}

\usepackage[framemethod=TikZ]{mdframed}

\mdfdefinestyle{MyFrame}{%
    linecolor=white,
    outerlinewidth=1pt,
    roundcorner=2pt,
    innertopmargin=\baselineskip,
    innerbottommargin=\baselineskip,
    innerrightmargin=5pt,
    innerleftmargin=5pt,
    backgroundcolor=black!5!white}

\newbool{APX}
\booltrue{APX}

\icmltitlerunning{Transformation Coding: Simple Objectives for Equivariant Representations}

\begin{document}

\twocolumn[
\icmltitle{Transformation Coding: Simple Objectives for Equivariant Representations}



\icmlsetsymbol{equal}{*}

\begin{icmlauthorlist}
\icmlauthor{Mehran Shakerinava}{equal,mcgill,mila}
\icmlauthor{Arnab Kumar Mondal}{equal,mcgill,mila}
\icmlauthor{Siamak Ravanbakhsh}{mcgill,mila}
\end{icmlauthorlist}

\icmlaffiliation{mcgill}{School of Computer Science, McGill University, Montr\'eal, Canada}
\icmlaffiliation{mila}{Mila- Quebec Artificial Intelligence Institute, Montr\'eal, Canada}

\icmlcorrespondingauthor{Arnab Kumar Mondal}{arnab.mondal@mila.quebec}
\icmlkeywords{Machine Learning, ICML}

\vskip 0.3in
]



\printAffiliationsAndNotice{\icmlEqualContribution} 

\begin{abstract}
We present a simple non-generative approach to deep representation learning that seeks equivariant deep embedding through simple objectives. In contrast to existing equivariant networks, our transformation coding approach does not constrain the choice of the feed-forward layer or the architecture and allows for an unknown group action on the input space. We introduce several such transformation coding objectives for different Lie groups such as the Euclidean, Orthogonal and the Unitary groups. When using product groups, the representation is decomposed and disentangled.
We show that the presence of additional information on different transformations improves disentanglement in transformation coding. We evaluate the representations learnt by transformation coding both qualitatively and quantitatively on downstream tasks, including reinforcement learning.
\end{abstract}

\section{Introduction}

Sample efficient representation learning is a key open challenge in deep learning for AI. Despite their success in many settings, generative representations are generally deemed sample inefficient. Therefore interest in self-supervision and contrastive learning techniques has significantly increased over the past few years. While contrastive learning objectives often create \emph{invariance} to certain (symmetry) transformations, in this paper, we pursue alternative objectives toward \emph{equivariant} representations. 

In contrast to previous equivariant networks, our approach, called \emph{transformation coding}, does not require any specialized architecture, and it can accommodate any non-linear transformation of the input. More importantly, one does not even need to know the correspondence between the transformations of the data and the underlying abstract group. For example, if our data consists of images before and after changing the camera angle, we do not need to know the actual angle. However, for example, we may need pairs of images in which the camera angle changes by the same amount.  

\begin{figure}[t!]
\begin{center}
  \centering
     \includegraphics[width=.8\linewidth]{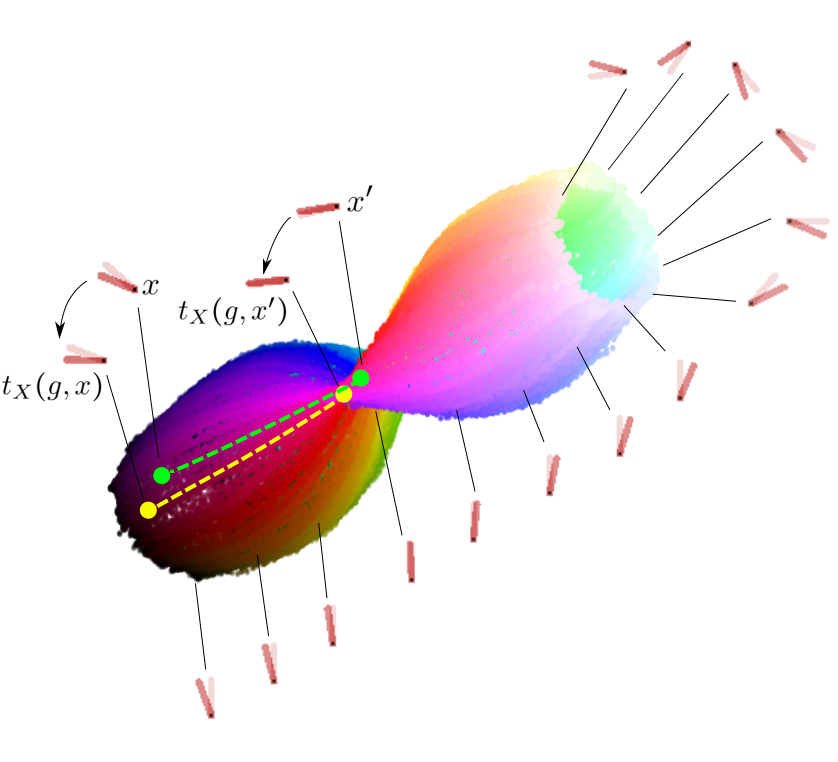}
     \caption{The $E(3)$-equivariant embedding for the pendulum. The input $x$ consists of a pair of images that together identify both the location and the velocity of a pendulum. The learned equivariant embedding encodes both the location (change of color) and velocity (change of brightness). The two circular ends (black and white) correspond to states of maximum velocity in opposite directions. The transformation coding objective for the \emph{Euclidean group} learns this embedding by preserving the pairwise distance between the codes before ($f(x), f(x')$) and after ($f(t_\Xs(\gg, x)), f(t_\Xs(\gg, x))$) transformations of the input by $t_\Xs$. For the pendulum, the transformations are in the form of positive or negative torque in some range. For yellow and green blobs on the manifold, the torque is applied clockwise.}
     \label{fig:pendulum}
     \end{center}
     \vskip -0.2in
\end{figure}

In the following, first, we observe that equivariance, in general, is a very weak inductive bias. In particular, we show that an injective code is equivariant to ``any'' transformation group. However, in this manifestation of equivariance, the group action on the code can be highly non-linear. Since the simplicity of the action on the latent space seems essential for equivariance to become a useful learning bias, we \emph{regularize} the group action on the code to make it ``simple''. This \emph{symmetry regularization} objective is group-dependent and the essence of our approach to transformation-based coding. 

Our symmetry objectives rely on the idea that the transformation groups preserve specific quantities. For example, the defining action of the Euclidean group preserves the Euclidean distance, and all Euclidean distance preserving transformations are of this form. Therefore, to enforce equivariance to the Euclidean group, it is sufficient to ensure that the embedding of any two data points has the same distance before and after the same transformation of the inputs; see~\cref{fig:pendulum}. We present similar objectives for several other groups.
Transformation coding combines this simple recipe with an injectivity constraint to produce equivariant representations without the need for a reconstruction loss or a hard 
constraint on the deep architecture.

\section{Related Works}\label{sec:related}
Finding effective priors and objectives for deep representation learning is an integral part of the quest for AI~\cite{bengio2013representation}. Recent years have witnessed a growing interest in self-supervision and, in particular, contrastive objectives~\cite{hadsell2006dimensionality,oord2018representation,chen2020simple,tian2019contrastive,he2020momentum,zbontar2021barlow,ermolov2021whitening}. 
While the use of transformations is prominent in these works, in many settings the objective encourages  \emph{invariance} to certain transformations, making such models useful for invariant downstream tasks such as classification. Similar to many of these methods, we also use transformed pairs to learn a representation, with the distinction of learning an \emph{equivariant} representation.

Learning both invariant and equivariant deep representations has been the subject of many works over the past decade. 
Many recent efforts in this direction have focused on the design of equivariant maps~\cite{wood1996representation,cohen2016group,ravanbakhsh2017equivariance,kondor2018generalization,cohen2018general,finzi2021practical,villar2021scalars} where the ``linear'' action of the group on the data is known. Due to this constraint, the application of these models has been focused on fixed geometric data such as images, sets, graphs, and spherical data, or physically motivated Poincare group, among others.

In the present work, the group action is unknown and possibly non-linear. Therefore, in contrast to many prior work, linear representation theory plays no significant role. Our setup is closer to the body of work on generative representation learning~\cite{burgess2018understanding,chen2016infogan, mita2021identifiable}, in which the (linear) transformation is applied to the latent space~\cite{quessard2020learning,worrall2017interpretable,kulkarni2015deep,lenc2016learning,cohen2014transformation,falorsi2018explorations}.
Among these generative coding methods, transforming autoencoder~\cite{hinton2011transforming} is a closely related early work that, in addition to equivariance, seeks to represent the part-whole hierarchy in the data. What additionally contrasts our work with the follow-up works on capsule networks~\cite{sabour2017dynamic,lenssen2018group} is that our transformation coding is agnostic to the choice of architecture and training. We only rely on our objective function to enforce equivariance.

When considering the Euclidean group, our equivariant code produces a manifold that preserves distances in the embedding space under non-linear transformations of the input. This embedding should not be confused with isometric embedding~\cite{tenenbaum2000global}, where the objective is to maintain the pairwise distances between points in the input and the embedding space.

\section{Background on Symmetry Transformations}\label{sec:background}
We can think of transformations as a set of bijective maps on a domain $\Xs$, and since these maps are composable, we can identify their compositional structure using a group $\Gg$. For this reason, such transformations are called {group actions}. To formally define transformation groups, we first define an abstract group.
A \emph{group} $\Gg$ is a set equipped with a binary operation, such that the set is closed under the operation $\gg \gg' \in \Gg \; \forall \gg, \gg' \in \Gg$, 
every $\gg \in \Gg$ has a unique inverse such that $\gg \gg^{-1} = \egr$, where $\egr$ is the identity element of the group, and the group operations are associative $(\gg \gg') \gg'' = \gg (\gg' \gg'')$.  

A \emph{\Gg-action} on a set $\Xs$ is defined by a function $t: \Gg \times \Xs \to \Xs$, which can be thought of as a bijective transformation parameterized by $\gg \in \Gg$.
In order to maintain the group structure, the action should satisfy the following two properties: 
(1) action of the identity is the identity transformation $t(\egr, x) = x$; (2) composition of two actions is equal to the action of the composition of group elements $t(\gg, t(\gg' x)) = t(\gg \gg', x)$. 
The action $t$ is \emph{faithful} to $\Gg$ if transformations of $X$ using each $\gg \in \Gg$ are unique -- \ie $\forall \gg, \gg' \; \exists x \in X \, \text{s.t.} \; t(\gg, x) \neq t(\gg', x)$.
If a $\Gg$-action is defined on a set $X$, we call $X$ a $\Gg$-set.

\section{Equivariance is Cheap, Actions Matter}\label{sec:cheap}
A symmetry-based representation or embedding is a function $f: \Xs \to \Zs$ such that both $\Xs$ and $\Zs$ are \Gg-sets, and furthermore $f$ ``knows about'' $\Gg$-actions, in the sense that transformations
of the input using $t_\Xs$ have the same effect as transformations of the output using some action $t_\Zs$:
\begin{align}\label{eq:def-equivariance}
    f(t_\Xs(g, x)) = t_\Zs(f(x), \gg) \quad \forall \gg, x \in \Gg \times \Xs
\end{align}

The following claim shows that despite many efforts in designing equivariant networks, simply asking for the representation to be equivariant is not a strong inductive bias, and we argue that the action matters.
\begin{proposition}
Given a transformation group $t_\Xs: \Gg \times \Xs \to \Xs$, the function $f: \Xs \to \Zs$ is an equivariant representation if $\forall \gg \in \Gg, x,x' \in \Xs$
\begin{equation}
\begin{split}\label{eq:claim-assumption}
    f(x) = f(x') \Leftrightarrow f(t_\Xs(\gg, x))& = f(t_\Xs(\gg, x')).
\end{split}    
\end{equation}
That is, two embeddings are identical iff they are identical for all transformations.
\end{proposition}

The proof is in the appendix.
The condition above is satisfied by all injective functions, indicating that many functions are equivariant to any group.

\begin{corollary}\label{cor:1}
Any \emph{injective} function $f:\Xs \to \Zs$ is equivariant to any transformation group $t_\Xs: \Gg \times \Xs \to \Xs$, if we define  $\Gg$ action on the embedding space as
\begin{align}\label{eq:ty2}
    t_\Zs(\gg, z) \doteq f(t_\Xs(\gg, f^{-1}(z))) \quad \forall \gg, z \in \Gg \times \Zs
\end{align}
\end{corollary}

The ramification of the results above in what follows is two-fold:\\
\textbf{1.} While injectivity ensures equivariance, 
the group action on the embedding as shown in~\cref{eq:ty2} can become highly non-linear. Intuitively, this action
recovers $x = f^{-1}(z)$, applies the group action $x' = t_\Xs(x)$ in the input domains and maps back to the embedding space $f(x')$ to ensure equivariance. In the following, we push $t_\Zs$ towards a simple linear \Gg-action through optimization of $f$. This objective can be interpreted as a \emph{symmetry regularization or a symmetry prior}.     

\textbf{2.} Although~\cref{cor:1} uses injectivity of $f$ for the entire $\Xs$, 
 we only need this for the data manifold. In practice, one could 
enforce injectivity on the training dataset $\Ds$ using loss functions defined on the training data, for example, using a hinge loss \cite{hadsell2006dimensionality}
\begin{align}
L_{\text{hinge}}(f, \Ds) & = \sum_{x,x'\neq x \in \Ds} \max \left (\epsilon - \norm{ f(x) - f(x'))}, 0 \right)
\end{align}
or other losses (barrier functions) that monotonically decrease with distance, such as $\frac{1}{\norm{f(x) - f(x')}}$, or its logarithm $-\log (\norm{f(x) - f(x')}) $. In experiments we use the logarithmic barrier function.

\begin{figure}
      \includegraphics[width=.9\linewidth,trim={10cm 3cm 19cm 0cm}, clip]{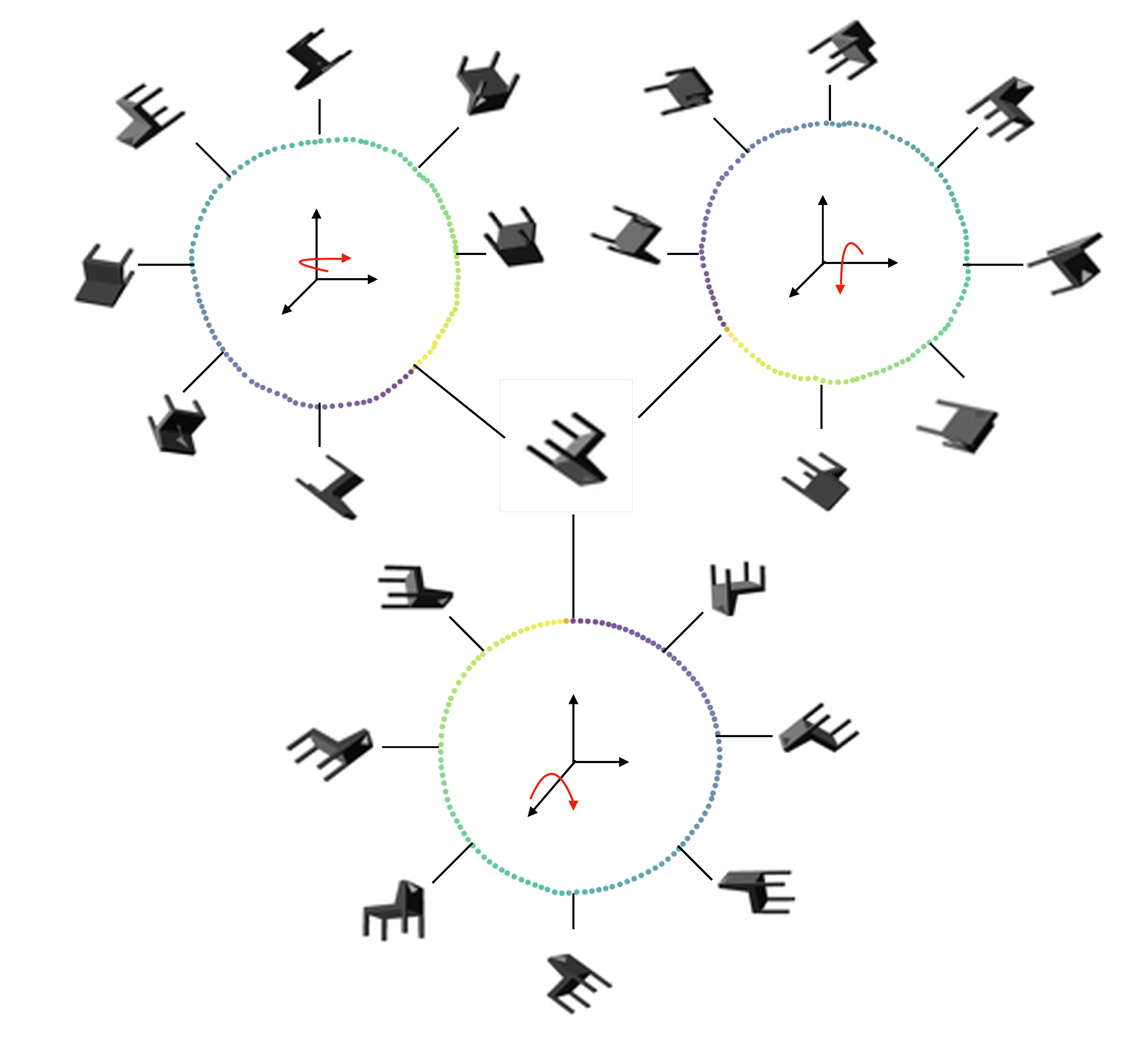}
      \caption{{Visualization of the latent projection for the rotating Chair dataset. The chair is rotated in three orthogonal axis from $0$ to $2\pi$. The latent embedding for each pose of the chair is projected from $16$D embedding space to a $2$D space for visualization. The colors of the representations is mapped to the the angle of rotation of the chair. We notice that the mapping function $f$ learnt is continuous with respect to the transformations of the object and it maps the rotations along an axis to a circular manifold. This is true for each orthogonal axis of rotation. 
  }}
     \label{fig:so3}
\end{figure}

\section{Symmetry Regularization Objectives}\label{sec:obj}
In learning equivariant representations, we often do not know the abstract group $\Gg$ and how it transforms our data, $t_\Xs$. We assume that one can pick a reasonable abstract group 
$\Gg$ that ``contains'' the real abstract group acting on the data -- \ie $\Gg$ action on the input may not be faithful. 
Our goal is to learn an $f: \Xs \to \Zs$ that is equivariant \wrt the actions $t_\Xs, t_\Zs$, where $t_\Xs: \Xs \times \Gg \to \Xs$ is unknown and $t_\Zs$ is some (simple) \Gg-action on $\Zs$ of our choosing. 

\subsection{More Informed but Less Practical Setting}
In the most informed case, the dataset also contains information about which group member $\gg \in \Gg$ can be used to transform $x$ to $x'$ -- that is the dataset consists of triples 
$\tuple{x, \gg, x_t = t_\Xs(\gg, x)}$. By having access to this information we can regularize the embedding using the following loss function: 
\begin{equation}
    L_\Gg^{\text{informed}}(f, D) = \sum_{\tuple{x,\gg, x_t} \in \Ds} \ell \big ( f(x_t) - t_\Zs(\gg, f(x) ) \big)
    \label{eq:informed}
\end{equation}
where $\ell$ is an appropriate loss function such as the square loss.
At its minimum, we have $f(x_t) = t_\Zs(\gg, f(x))$ or $f(t_\Xs(\gg, x)) = t_\Zs(\gg, f(x))$, enforcing equivariance condition of~\cref{eq:def-equivariance}. However, even if the optimal value is not reach, due to its injectivity, $f$ is still $\Gg$-equivariant, and the the objective above is regularizing or simplifying the $\Gg$ action on the code.

The assumption of having access to $\gg$ is often not realistic.
As an example, in the Reinforcement Learning (RL) settings, where we have access to the action of the agent that can \emph{transform} the state, although we have a triplet $\tuple{x, a, x_t}$, often $a$ cannot be trivially mapped to a group element.
Fortunately, using the fact that certain group actions are symmetries of some structures, we may still learn an equivariant embedding, even if we do not have the group information tied to the dataset. In this approach, the loss
function and other potential sources of information in $\Ds$ will be group-specific. As we see shortly, for finite groups, we a symmetry regularization without any additional information is viable. However, for the remaining groups discussed in this paper, we need to be able to identify pairs of identical transformations -- \ie transformations that use the ``same'' group member -- although we do not need to know the actual group member. 

\subsection{Finite Groups}
Given two instances $x,x' \in \Ds$, we may ``optimize'' for the choice of $\gg \in \Gg$, such that
$t_\Zs(f(x), \gg) \approx f(x')$, changing~\cref{eq:informed} to  
\begin{align}\label{eq:perm}
    L_{\Gg}(f, \Ds) = \sum_{x,x' \in \Ds} \, \min_{\gg \in \Gg} \, \ell \,\big ( f(x) - t_\Zs \big(\gg, f(x')\big) \big ).
\end{align}
This approach 
may be suitable for finite groups, where one can enumerate $\gg \in \Gg$ to perform the minimization inside the outer optimization.
Since any finite group has a permutation representation, we can always define $t_\Zs$ to permute blocks of the latent representation. In this case, the loss function above involves a search to find the best permutation within a group that matches the embeddings before and after the transformation.

For some permutation groups, this search can be performed more efficiently. 
For example, in the case of the symmetric group, the loss function of~\cref{eq:perm} is also sometimes known as earth mover's distance~\cite{fan2017point}.
One can also use the Chamfer distance
to approximate this loss function.
In contrast to the existing permutation equivariant networks~\cite{zaheer2017deep,qi2017pointnet}, here the permutation action on the input is not limited to the permutation of a known group of variables -- that is, the representation could be equivariant to shuffling of a priori unknown entities/objects in the input.  
\subsection{Euclidean Group}
The defining action of the Euclidean group $E(n)$ is the set of transformations that preserve the Euclidean distance between any two points in $\Reals^n$, \aka isometries. These transformations are compositions of translations, rotations, and reflections.
Since for the real domain, all isometries are linear and belong to $E(n)$,
we can enforce the group structure on the embedding by ensuring that distances 
between the embeddings before and after any transformation match.
For this, we need the dataset $\Ds$ to be a set of pairs of pairs $\btuple{\tuple{x,x_t = t_\Xs(\gg, x)},\tuple{x', x'_t = t_\Xs(\gg, x')}}$, where $x,x'$ are transformed 
using the same \emph{unknown} group member $\gg$. 
Distance-preservation loss below combined with injection loss are sufficient to produce an $E(n)$-regularized embedding:
\begin{align}
    L_{E(n)}(f, \Ds) =  \smashoperator{\sum_{\substack{ \big(\tuple{x,x_t}, \tuple{x',x'_t} \big) \in \Ds}}} 
        \ell \big ( \overbrace{\norm{ f(x) - f(x') }}^{\substack{\text{distance before the }\\ \text{transformation}}} - \overbrace{\norm{f(x_t) - f(x'_t)}}^{\substack{\text{distance after the} \\ \text{transformation}}}  \big ) \label{eq:en}
\end{align}

In the standard RL setup, where we have access to triplets $(s, a, s')$, we can easily form $D$ by unrolling an episode and collecting two different state transitions corresponding to a particular action. In practice, with a finite actions, we can efficiently generate this dataset by keeping separate buffer for each action where we store state transitions for that action and sample from that buffer to train the embedding function $f$. 

\subsection{Orthogonal and Unitary Groups }
\label{subsec:ortho}
The defining action of the orthogonal group $O(n)$ preserves the inner product between two vectors; these are exactly Euclidean isometries that fix the origin. The analogous group in the complex domain is the unitary group, which preserves the complex inner product. 
Our symmetry-regularization objective penalizes the change in the inner product of two embeddings before and after the same transformation:
\begin{align*}
    \label{eq:on}
    L_{O(n)}(f, \Ds) = \quad \smashoperator{\sum_{\substack{ \big(\tuple{x,x_t}, \tuple{x',x'_t} \big) \in \Ds}}}  \qquad \, \ell\, \big (\, { f(x)^\top f(x') } - f(x_t)^\top f(x'_t) \, \big ).
\end{align*}

For the unitary group one additionally needs to embed to complex domain $\Zs = \Complex^n$ or use separate real and imaginary codes use the inner product of the complex domain.

\begin{figure}
  \begin{center}
    \includegraphics[width=0.2\textwidth]{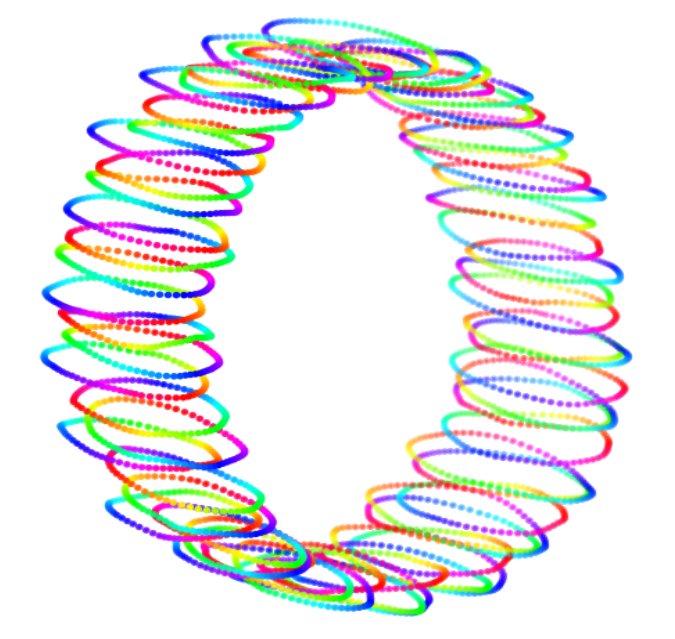}  
    \hfill
    \includegraphics[width=0.2\textwidth,trim={50 70 50 50},clip]{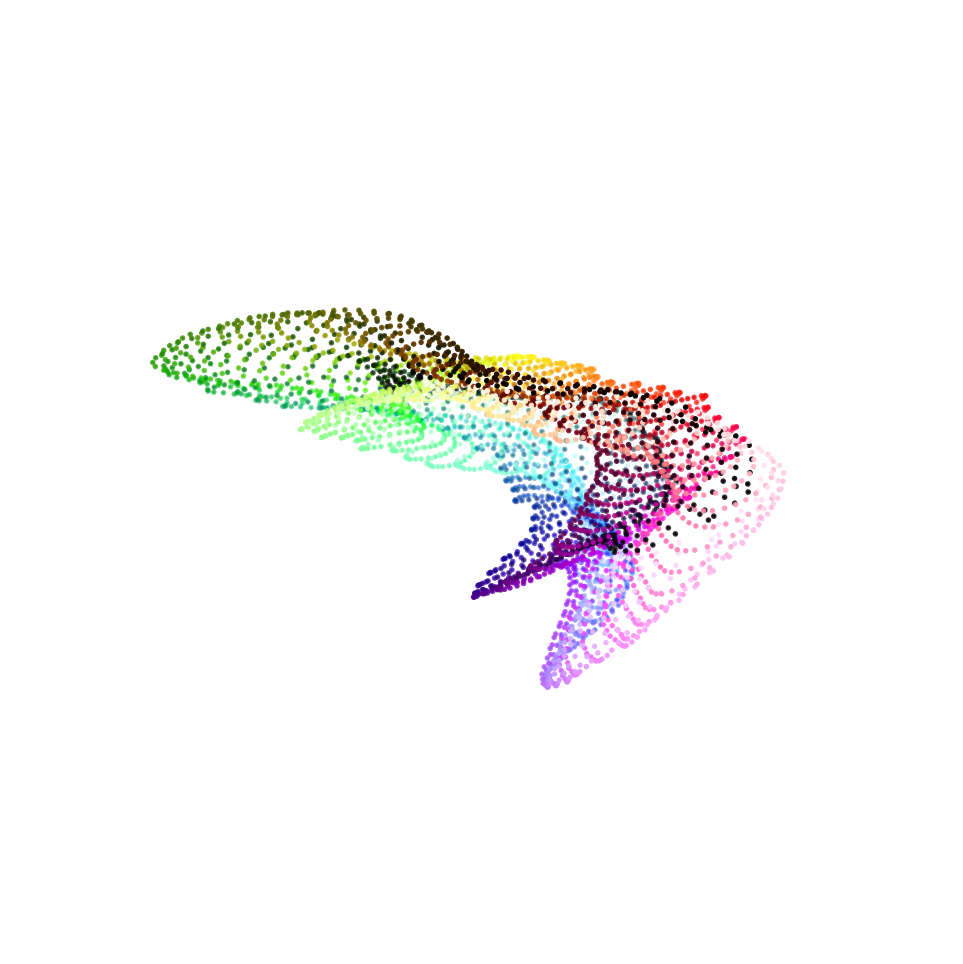}
  \end{center}
    \vspace*{-1em}
  \caption{{(left) Conformal vs. (right) VAE embedding  of double-bump world.}}
  \label{fig:conformal}
  \vspace*{-1em}
\end{figure}

\subsection{Conformal Group}
Conformal group is the group of transformations that preserve the angle.
In a Euclidean embedding, these transformations include a combination of translation, rotation, dilation, and inversion with respect to an $n-1$-sphere. 
To enforce this group structure we need a triplets of inputs, before and after a transformation $\btuple{\tuple{x,x_t},\tuple{x', x'_t}, \tuple{x'', x''_t}}$, so that we can calculate the angle in the embedding. Similar to distance-preserving loss of~\cref{eq:en}, this objective enforces an \emph{invariance}; this time we enforce the invariance of the angle between the triplet, before and the transformation:

\begin{equation}
\begin{split}
  \label{eq:conformal}
    L_{CO(n)}(f, \Ds) =&\quad \smashoperator{\sum_{\substack{ \big( \tuple{x,x_t},\tuple{x', x'_t}, \tuple{x'', x''_t} \big) \in \Ds}}} \quad
    \ell\, \big (\, \cos \big(\angle f(x),f(x'), f(x'') \big) \\ & - \cos (\angle f(x'_t),f(x'_t), f(x''_t) \big)  
\end{split}
\end{equation}
where $ \cos\big(\angle y, y', y'' \big) = \frac{(y - y')^\top (y'' - y')}{\norm{y - y'}\norm{y'' - y'}}$ is the cosine of the angle between the  three embeddings. 

This objective imposes a weaker constraint on the embedding than the distance preservation of the Euclidean group -- preserving Euclidean distance implies preserving the angle. Moreover, it has an additional benefit that compared to $L_{E(n)}$ the loss cannot be minimized by simply shrinking the embedding, therefore in practice, the injection enforcing losses of~\cref{sec:cheap} is no longer necessary when using conformal symmetry regularization.

\begin{figure*}[t!]
  \centering
    \includegraphics[width=0.8\linewidth]{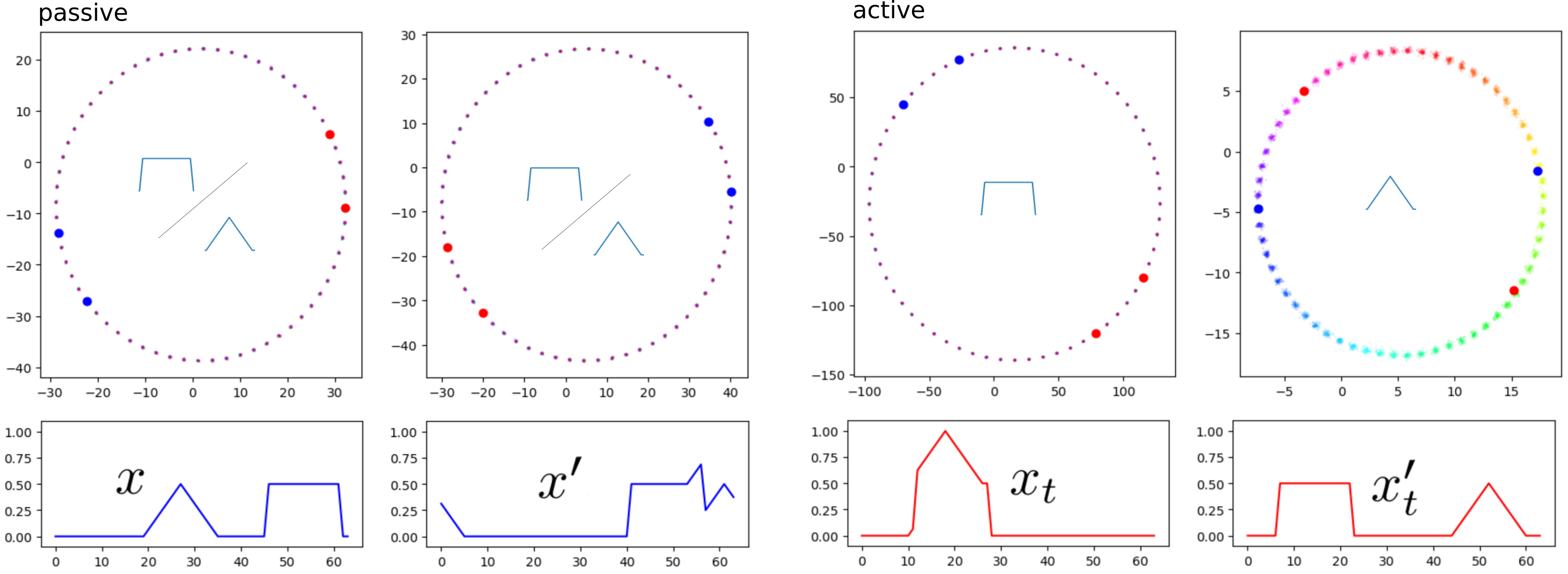}
  \vspace{-0.7 \baselineskip}
  \caption{{Active versus passive decomposition of the embedding for the double-bump world into $E(2) \times E(2)$-set. In active decomposition (right), one of the manifolds encodes the circular translation of the triangle bump, while the second one represents the location of the square bump. Various colors indicate the location of the triangle.
  In the case of passive decomposition (left), since the decomposition is not guided by the transformation of individual shapes, 
  the manifolds jointly encode the location of each bump type.
  The figure shows the embedding for two inputs before, $(x,x')$ in blue, and after, $(x_t, x'_t)$ in red, the \emph{same} transformation. This transformation cyclically shifts both the triangle and the square to the left, but the amount of translation is larger for the square. In both passive and active decomposition, the Euclidean distance is preserved by the transformation -- the red points have the same distance with each other as the blue points on every manifold.
  }}
  \label{fig:decomposition}
\end{figure*}

\section{Decomposing the Representation}
\citet{higgins2018towards} suggested a notion of disentangled representation based on decomposition of the abstract group into a direct product form $\Gg = \Gg_1 \times \ldots \times \Gg_k$. There are two approaches to learning such decomposed representation in transformation coding, depending on whether or not we can perform certain types of transformations in isolation. For example, an RL agent may transform its environment through actions like moving a single limb that can be performed in isolation. In this case, we call the decomposition \emph{active} to contrast it with the \emph{passive} case where the action of different subgroups is always mixed in our dataset.

\subsection{Active Decomposition}
Let $\Gg = \{(\gg_1, \ldots, \gg_k) \in \Gg_1 \times \ldots \times \Gg_k\}$, where  
$$\Gg_i \cong \{(\overbrace{e, \ldots, e,}^{i-1}\gg_i, e, \ldots, e) \in \Gg\}$$
can be identified with a normal subgroup of $\Gg$. 
In active decomposition, sub-groups can act in isolation and therefore we have
$k$ types of tuples in our dataset $\Ds_1,\ldots,\Ds_k \subset \Ds$. Each subset $D_i$ is associated with actions of a subgroup $\Gg_i$ using $t_\Xs(({e, \ldots, e,} \gg_i, e \ldots, e), \cdot))$, $\gg_i \in \Gg_i$. 

In this setting, the representation $f:\Xs \to \Zs = \Zs_1 \times \ldots \times \Zs_k$ can be thought of as $k$ separate functions where $f_i: \Xs\to \Zs_i$ is equivariant to $\Gg_i$-action and invariant to all $\Gg_j, j\neq i$ actions.
This gives the following objective 
\begin{align}\label{eq:product}
    L^{\text{active}}_{\Gg}(f, \Ds) = \sum_{i=1}^{k} {\underbrace{L_{\Gg_i}(f_i, \Ds_i)}_{\text{equivariance to $\Gg_i$}} + \underbrace{L^{inv.}_{\Gg / \Gg_i}(f_i, \Ds \backslash D_i)}_{\text{invariance to $\Gg_j$ for } j \neq i}}
\end{align}
where  $L^{inv.}_\Gg(f, \Ds)$ enforces invariance of $f$ to $\Gg$-transformations in $\Ds$ -- e.g., by penalizing $\norm{f(x) - f(t_\Xs(\gg, x))}$.
In addition to  isolated actions assumed in this setting we may have mixed actions.
Moreover, if this mixing involves a known sparse subset of sub-groups (\eg a subset of joints) the loss function of~\cref{eq:product} can be modified accordingly.

\subsection{Passive Decomposition}\label{sec:passive}
When we have no control over transformations, and we are simply given the data, it is still possible to use an abstract group that has a product form. Here again, $f:\Xs \to \Zs = \Zs_1 \times \ldots \times \Zs_k$, but the loss function is simply enforced on each block separately -- \ie
$L^{\text{passive}}_{\Gg}(f, D) = \sum_{i=1}^{k} L_{\Gg_i}(f_i, D)$, where $L_{\Gg_i}(f_i, D)$ is a transformation coding objective from~\cref{sec:obj}.

\section{Experiments}\label{sec:experiments}
We conducted many experiments to qualitatively study the representation learned by transformation coding, its ability to produce a
disentangled representation, and quantitatively compare against simple baselines in both representation learning and downstream RL tasks.
For details of architecture and training, see \cref{appdx:Implementation}.

\begin{wrapfigure}{r}{0.4\linewidth}
\vspace{-5em}
  \begin{center}
    \includegraphics[width=\linewidth,trim={70 100 50 50},clip]{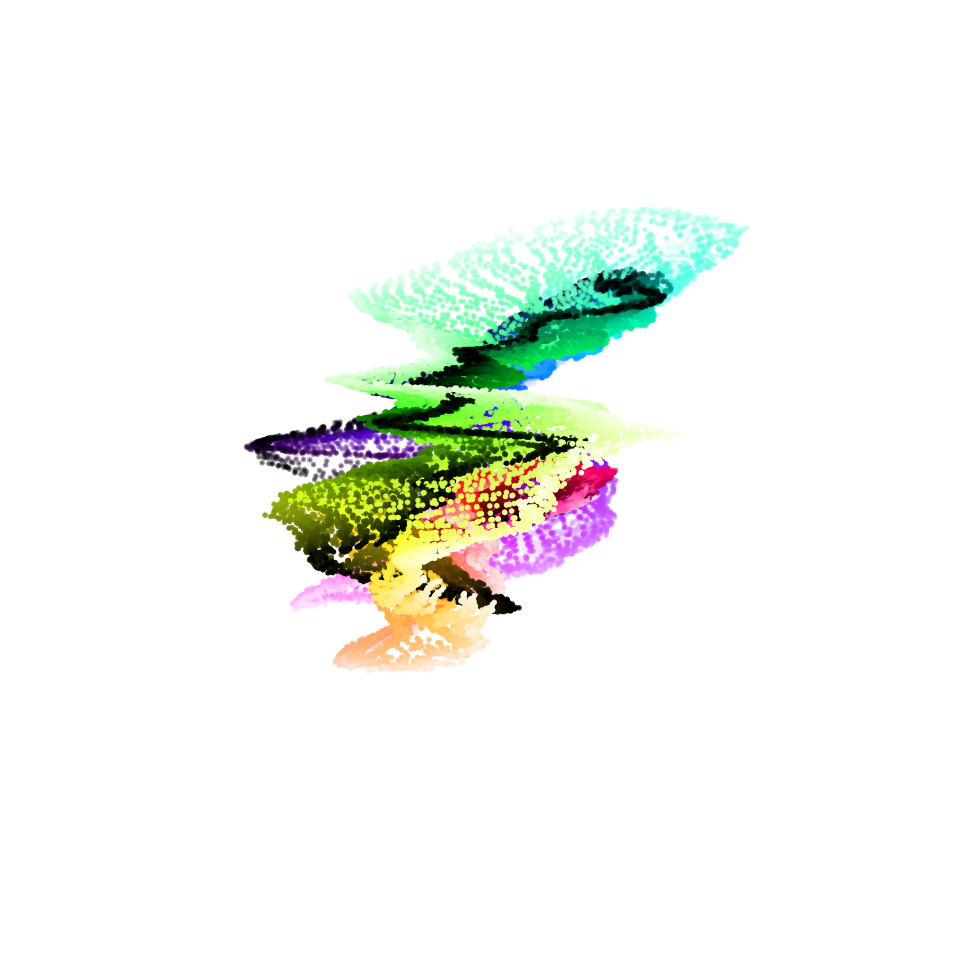}
  \end{center}
  \vspace*{-1em}
  \caption{VAE embedding for the pendulum example; compare againts \cref{fig:pendulum}}\label{fig:vae-pendulum}
\end{wrapfigure}

\subsection{Qualitative Analysis}
In this section, we visualize the representation learned for two examples from the Gym environment \citep{brockman2016openai}, including the pendulum and the mountain car (see \cref{appdx:MountainCar}, followed by an experiment involving a rotating object where we know the ideal embedding in the form of $SO(3)$ manifold.
Finally, \cref{fig:conformal} visualizes a conformal embedding for double-bump world. In most cases we also visualize Variational AutoEncoder (VAE) \cite{kingma2013auto} embeddings for comparison.
Our objective here is to visually demonstrate the behavior of transformation coding and its remarkable ability to learn embeddings that are informed by non-linear transformations of the input. 

\subsubsection{The Pendulum}
\label{exp:pendulum}
For this experiment, the input $x$ is two consecutive frames of the pendulum that have been grayscaled and downsampled to 32×32 pixels. The action-space is a range of torques that can be applied on the base of the pendulum. We use the action to transform the data. We use the objective of~\cref{eq:en} to learn an $E(3)$-equivariant representation. To efficiently estimate $L_{E(n)}$, we use a mini-batch that consists of 64 randomly sampled observations from the environment and their transformations via three randomly sampled actions ($4 \times 64$ samples in total). Once the embedding $f(x)$ is produced, the pairwise distance between all $64^2$ pairs is calculated, and the mini-batch loss penalizes the change in these distances between any pair of transformations in the mini-batch. Therefore, using this mini-batching procedure, the complexity of calculating the loss grows quadratically with both the number of independent samples and transformations.

The model on itself learns to parameterize the embedding using the location and the velocity of the pendulum from the input data; see \cref{fig:pendulum}.\footnote{A natural parametrization of a slow-moving pendulum using location and velocity is a cylinder, where the $SO(2)$ encodes the location, and $\Reals$ encodes the velocity. At high velocities, the two ends of the cylinder twist and meet, forming a Klein bottle.
However, the learned manifold of~\cref{fig:pendulum} is different from both; it has a twist in the middle. This twist turns out to be necessary for enforcing $E(3)$-equivariance. Because of the twist, the movement of the pendulum has the same direction around the circles on both sides. This is necessary for maintaining the distance of two points on opposite sides of the manifold after a transformation.} For comparison \cref{fig:vae-pendulum} shows the embedding learned using a 
VAE from the same dataset. We perform a similar experiment in Moutain Car Environment and report its results in \cref{appdx:MountainCar}. 

\subsubsection{Rotating Chair}
\label{exp:rotating_chair}
We consider a 3D chair from ModelNet40 \cite{wu20153d} and transform it through the action of the group $SO(3)$.  The group action in the input space is given by 2D projection into a $48\times48$ image after 3D rotation of the chair. 
While the group of interest is $SO(3)$, we use Euclidean regularization loss of \cref{eq:en}. In choosing the abstract group, we often only need to ensure that the group is large enough to contain the ground truth as a subgroup. 
Although this results in a stronger symmetry regularization on the embedding, a $G$-equivariant embedding is equivariant to any $H \leq G$.
This means that, for example, an $E(2)$ equivariant embedding can be useful for its finite subgroups such as dihedral or cyclic groups.
We embed the chair in $\Reals^{16}$ using transformation coding\footnote{Note that while $SO(3)$ manifold is 3-dimensional, its isometric embedding requires a higher number of dimensions. Using a larger embedding dimension also often helps with the optimization of our symmetry regularization loss.}
 and visualize the latent by rotating the chair along three orthogonal axes and projecting the latent codes into a 2D space. \cref{fig:so3} shows three circular latent traversals corresponding to rotation around each axis, which is consistent with the structure of $SO(3)$ manifold. The process of learning $SO(3)$ manifold is a challenging task, and previous works assumed that the group member corresponding to each transformation is given  \cite{quessard2020learning, anonymous2022learning}. In contrast, we only use the observations corresponding to similar actions during training and not the group members themselves. As we see later, this is critical in settings such as RL, where group information is unavailable. We were not able to produce a similar latent traversal for VAE due to collapse when rotating around some axes. 

\subsubsection{Conformal Group for Double-Bump World}
\label{exp:conformal_bumpworld}
The double-bump world consists of a rectangular bump signal and a triangular bump signal, both of which have been cyclically shifted and superimposed. Any transformation in this dataset can be denoted as a pair $(\Delta_1, \Delta_2)$ which cyclically shifts the rectangular bump by $\Delta_1$ and the triangular bump by $\Delta_2$. In our experiments, the length of the signal is 64, and the length of the bump is 16. We used conformal loss of~\cref{eq:conformal} in this experiment. \cref{fig:conformal}(left) shows the random project of the embedding, where the colors change as the triangle bump moves. The figure suggests that transformation coding is able to successfully learn to represent a data point using the location of two bumps.

\subsection{Experiments on Active and Passive Decomposition}\label{sec:exp-decomposition}
In this section, we first contrast active and passive decomposition in their ability to disentangle the two bumps in the double bump world.
We observe that while both can decompose the embedding into a product form $SO(2) \times SO(2)$, only active decomposition leads to disentanglement.
\cref{sec:exp-ego} applies active decomposition to a more complex setting of ego-motion, where transformation coding can decompose the 
representation of the agent's state into location and orientation.

\subsubsection{Decomposition of the Double-Bump World}
\label{exp:decom_bump world}
Next, we compare the active and passive decomposition for the same double-bump world. While the ground truth is $SO(2) \times SO(2)$, the method uses the larger group $E(2) \times E(2)$. 
In the active case, each subgroup moves one of the bumps, the loss of~\cref{eq:product} is used to learn an embedding for each subgroup. In the passive case, both bumps move randomly.

\cref{fig:decomposition} compares the decomposed embedding found in each case.
While in both cases, the $SO(2) \times SO(2)$ torus is decomposed into a product of
circles, only the active case successfully disentangles the two bumps. Note that the color of each point is based on the location of the triangle bump. Our results agree with  \citet{caselles2019symmetry} who claim that learning a disentangled representation requires interaction with the environment; see also \cite{painter2020linear,mita2021identifiable}. However, we note that while the disentangling of the bump movements do not happen in the passive case, we can still successfully ``decompose'' the embedding.

\begin{figure}[ht!]
  \centering
    \includegraphics[width=0.9\linewidth]{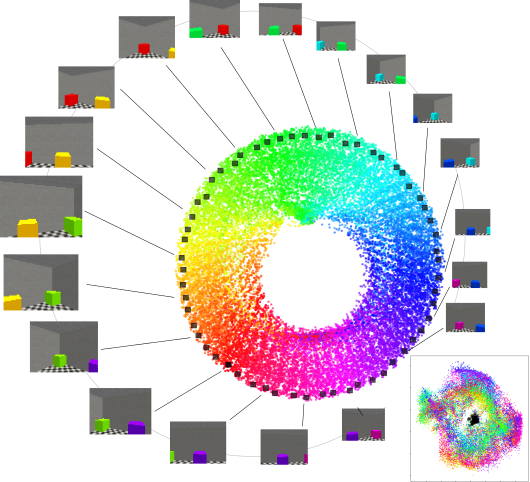}
  \caption{
    Decomposition of the ego-motion manifold using $E(2) \times E(2)$ equivariant coding. 
    The dataset contains a first-person view of a room. 
  Transformations include right-left rotation and forward-backward movement. The equivariant embedding is produced by active decomposition using these two transformations, where the ring-structured manifold corresponds to the rotation action, and the smaller manifold corresponds to translations. Color-coding shows the true angle of the image. The black square markers show the traversal of the embedding as the agent rotates while standing in the middle of the room. Note that in the second manifold,  black squares are concentrated in the center. }
  \label{fig:maze}
\end{figure}


\subsubsection{Active Decomposition for Ego-Motion}\label{sec:exp-ego}
\begin{wrapfigure}{r}{0.25\linewidth}
  \vspace*{-1em}
  \begin{center}
    \includegraphics[width=0.8\linewidth]{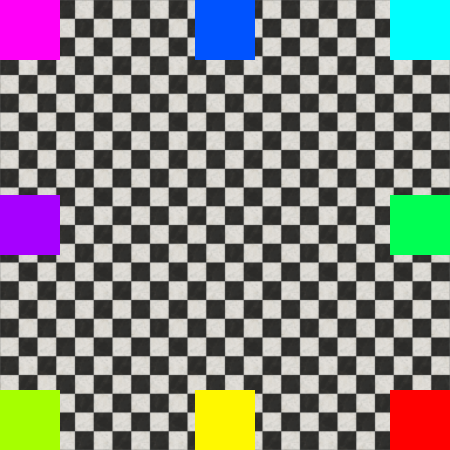}
  \end{center}
  \vspace*{-1em}
  \caption{\footnotesize{Top view of the room}}\label{fig:maze_map}
\end{wrapfigure}
We used a modified version of the single-room environment of MiniWorld~\cite{gym_miniworld} for this experiment.
The agent is standing in a 3D room containing eight differently colored boxes around the walls. A map of the room can be seen in~\cref{fig:maze_map}. An observation consists of a first-person view of the room, downsampled to $32 \times 32$ pixels. The agent can rotate left/right or move forward/backward. We learn an $E(2) \times E(2)$ equivariant embedding using the active decomposition objective of~\cref{eq:product}.
Each mini-batch consists of 64 random observations and the result of applying all four actions in those states ($4 \times 64$ samples in total).

\cref{fig:maze} visualizes the embedding of the input in two sub-figures, where the more prominent figure shows the embedding corresponding to the rotation action, and the more petite figure (bottom right) shows the embedding corresponding to forward-backward movement. The first figure also shows the first-person view when the agent rotates while standing in the middle of the room. The corresponding markers collapse around the center of the second embedding, demonstrating an intuitive embedding, parameterized by rotation angle and location. 
Walking straight across the room also produces the expected behavior of traversing the second manifold while the rotation angle, for the most part, remains fixed (not shown).

\subsection{Quantitave Evaluation in Downstream Tasks}
This section quantitatively shows the effectiveness of transformation coding for World Modelling and RL. 

\subsubsection{World Modelling}
We select Atari games Pong and Space Invader for the world modeling experiments as our environment. These environments are previously used by \cite{Kipf2020Contrastive} to evaluate their model, Contrastive Structured World Model (C-SWM). 
We train the encoder using our Euclidean transformation coding objective, freeze it, and then learn a Multi-Layer Perceptron (MLP) based transition function in the latent space. Our evaluation scheme follows \cite{Kipf2020Contrastive}. We report Hits at Rank 1 (H@1) and Mean Reciprocal Rank (MRR), which are invariant to the embedding scale. These evaluation metrics measure the relative closeness of the next state's representation predicted by the transition model and the representation of the observed next state. To measure the relative closeness, we use a set of reference state representations (embedding of random observations from the experience buffer). \cref{tab:worldmodel} reports these measures and show that a simple transition model learned on top of our embedding outperforms C-SWM in both games. Other reported baselines use an AutoEncodcer (AE) and a Variational AutoEncoder (VAE) to learn embeddings. 

\begin{table}[t]
\caption{Hits at Rank 1 (H@1) and Mean Reciprocal Rank (MRR) of different method. We report our models performance over 5 random seeds for Pong and Space Invaders.}\label{tab:worldmodel}
\begin{center}
\begin{small}
\begin{sc}
\begin{adjustbox}{width=0.47\textwidth}
\begin{tabular}{llll}
\toprule
 Environment & Method & H@1 & MRR \\
\midrule
\multirow{4}{*}{Atari Pong} & World Model(AE)   & 23.8\scalebox{0.65}{$\pm$ 3.3} & 44.7\scalebox{0.65}{$\pm$ 2.4} \\
&World Model(VAE) & 1.0\scalebox{0.65}{$\pm$ 0.0} & 5.1\scalebox{0.65}{$\pm$ 0.1}\\
&C-SWM     & 36.5\scalebox{0.65}{$\pm$ 5.6} & 56.2\scalebox{0.65}{$\pm$ 6.2} \\
&\textbf{Ours}  & \textbf{45.2\scalebox{0.65}{$\pm$ 3.4}} & \textbf{60.2\scalebox{0.65}{$\pm$ 3.9}}\\
\midrule
\multirow{4}{*}{Space Invaders} & World Model(AE)   & 40.2\scalebox{0.65}{$\pm$ 3.3} & 59.6\scalebox{0.65}{$\pm$ 3.5} \\
&World Model(VAE) & 1.0\scalebox{0.65}{$\pm$5.3} & 5.3\scalebox{0.65}{$\pm$ 0.1}\\
&C-SWM     & 48.5\scalebox{0.65}{$\pm$ 7.0} & 66.1\scalebox{0.65}{$\pm$ 6.6} \\
&\textbf{Ours}  & \textbf{54.2\scalebox{0.65}{$\pm$ 6.3}} & \textbf{68.7\scalebox{0.65}{$\pm$ 5.1}}\\
\bottomrule
\end{tabular}
\end{adjustbox}
\end{sc}
\end{small}
\end{center}
\end{table}

\subsubsection{Reinforcement Learning}
Next, we consider three Mujoco environments: InvertedPendulum, Reacher, and Swimmer from OpenAI Gym \cite{brockman2016openai}. We introduce two variations of our model to test the usefulness of transformation coding in the context of RL. The first variation uses a fixed encoder that is pretrained using transformation coding and then trains an MLP heads for function approximations in the downstream RL algorithm (TC-decoupled); that is, the low-dimensional embedding is used as a substitute for the high-dimensional input data without further adjustment. The second variation allows for finetuning during the reinforcement learning stage (TC-finetuned). We designed two other baseline models using AutoEncoder (AE) to train the encoder and call them AE-decoupled and AE-finetuned. We use Proximal Policy Optimization (PPO) \cite{schulman2017proximal} as the underlying RL algorithm. To evaluate the data-efficiency of these models, we report the average reward collected over 10 episodes in the first 100,000 steps for Reacher and Swimmer and 30,000 steps for Inverted Pendulum in \cref{tab:rl} (since Inverted Pendulum generally learns faster, we took a fewer number of steps.) 

We see that learned representations adequately capture the structure of the environment in Inverted Pendulum since the RL agent just trained on the fixed representation (TC-decoupled) outperforms vanilla PPO. 
In contrast, AE does not seem to capture the structure, at least not in a form suitable for learning a policy. In Reacher, TC-decoupled performs poorly compared to AE-decoupled. We believe that this is because the representation is focused on transformations due to the agent's actions, and details that can be valuable from the reward's perspective can be ignored -- in this case, the small object that the Reacher should reach. This observation points to a significant limitation of our approach, that is in this case resolved by finetuning. Like Reacher, we see that learning the agent's transformations is not enough to get all the reward information as the background movement decides how far the agent has swum. Indeed, allowing the encoder to finetune allows the representations to reflect the reward information and improve performance. 

\begin{table}[t]
\caption{Average reward collected over 10 episodes for various models in Inverted Pendulum, Reacher and Swimmer. We provide the standard error for each of them using 5 different random seeds for each experiment.}
\label{tab:rl}
\begin{center}
\begin{small}
\begin{sc}
\begin{adjustbox}{width=0.47\textwidth}
\begin{tabular}{lccc}
\toprule
Methods & InvertedPendulum & Reacher & Swimmer \\
\midrule
 Vanilla & 500\scalebox{0.65}{$\pm$ 150}   & -11\scalebox{0.65}{$\pm$ 2.5} & 25.6\scalebox{0.65}{$\pm$ 3.4} \\
 AE-decoupled & 30\scalebox{0.65}{$\pm$ 15} & -13\scalebox{0.65}{$\pm$ 3.0} & 16\scalebox{0.65}{$\pm$ 3.9}\\
 AE-finetuned & 580\scalebox{0.65}{$\pm$ 130} & -11.5\scalebox{0.65}{$\pm$ 3.2} & 26\scalebox{0.65}{$\pm$ 4.3} \\
 TC-decoupled & 800\scalebox{0.65}{$\pm$ 180}  & -14.5\scalebox{0.65}{$\pm$ 3.1} & 21\scalebox{0.65}{$\pm$ 4.1}\\
 TC-finetuned &\textbf{950\scalebox{0.65}{$\pm$ 50}}  & \textbf{-10\scalebox{0.65}{$\pm$ 3.4}} & \textbf{31.5\scalebox{0.65}{$\pm$ 3.9}}\\
\bottomrule
\end{tabular}
\end{adjustbox}
\end{sc}
\end{small}
\end{center}
\end{table}

\section*{Conclusion}
Transformation coding is shown to be a simple, intuitive, and yet effective approach for learning equivariant representations, where the group action on the input is potentially non-linear and unknown. The idea is to learn a bijection that is regularized towards simple actions of the abstract group on the latent space. 
Symmetry regularization loss is specific to each group and relies on the preservation of certain quantities. This is reminiscent of the
unification of geometries using group theory in Klein's Erlangen program; preservation of distance, inner product, and angle are used to define
latent representations equivariant to Euclidean, orthonormal, and conformal groups. This also points to an important limitation:
 a recipe for creating symmetry regularization losses for general Lie groups is currently missing. 
While this paper showcases the effectiveness of transformation coding in several qualitative and quantitative experiments, we see many avenues 
for further exploration in future works.

\nocite{langley00}

\bibliography{example_paper}
\bibliographystyle{icml2022}

\newpage
\appendix
\onecolumn
\section{Proof of Claims}

\begin{proof} of Claim 1 \\
Let $\sim_f$ be an equivalence relation on $\Xs$, such that two points are ``equivalent'' if they have the same embedding 
$x \sim x' \Leftrightarrow f(x) = f(x')$.
We use $[x]_\sim$ to denote the equivalence class of $x$. 
To get an intuition for this result, first consider an injective $f$, where the equivalence classes are trivial $[x]_\sim = x$.
In this case for any $\Gg$-set $\Xs$, $f$ is $\Gg$-equivariant with $\Gg$-action on $\Zs$ defined by 
\begin{align}\label{eq:ty}
    t_\Zs(\gg, y) \doteq f(t_\Xs(\gg, f^{-1}(y))) \quad \forall \gg, y \in \Gg \times \Zs
\end{align}
Now to see why $f$ is equivariant to any action $t_\Xs$ and the corresponding $t_\Zs$ defined above, simply
replace the definition of $t_\Zs$ into definition of equivariance \cref{eq:def-equivariance}
\begin{align}
    t_\Zs(\gg, f(x)) = f(t_\Xs(\gg, f^{-1}(f(x)))) = f(t_\Xs(\gg, x))
\end{align}

For general functions, note that $f^{-1}(f(x)) = [x]_\sim$. 
The equation above makes sense iff $t_\Xs(\gg, x') = t_\Xs(\gg, x'') \forall x',x'' \in [x]_\sim$, which is basically the assumption of \cref{eq:claim-assumption}. This means
$[t_\Xs(\gg, x)]_\sim \defeq [t_\Xs(\gg, x)]_\sim$, and using the $t_\Zs$ of \cref{eq:ty} in the definition  
of equivariance, we see that its condition is satisfied
\begin{align}
    t_\Zs(\gg, f(x)) = f(t_\Xs(\gg, f^{-1}(f(x)))) = f(t_\Xs(\gg, [x]_\sim)) =  f([t_\Xs(\gg, x)]_\sim) = f(t_\Xs(\gg, x))
\end{align}
\end{proof}
\begin{figure*}[ht]
  \centering
  \begin{minipage}[c]{.35\linewidth}
    \includegraphics[width=.7\linewidth]{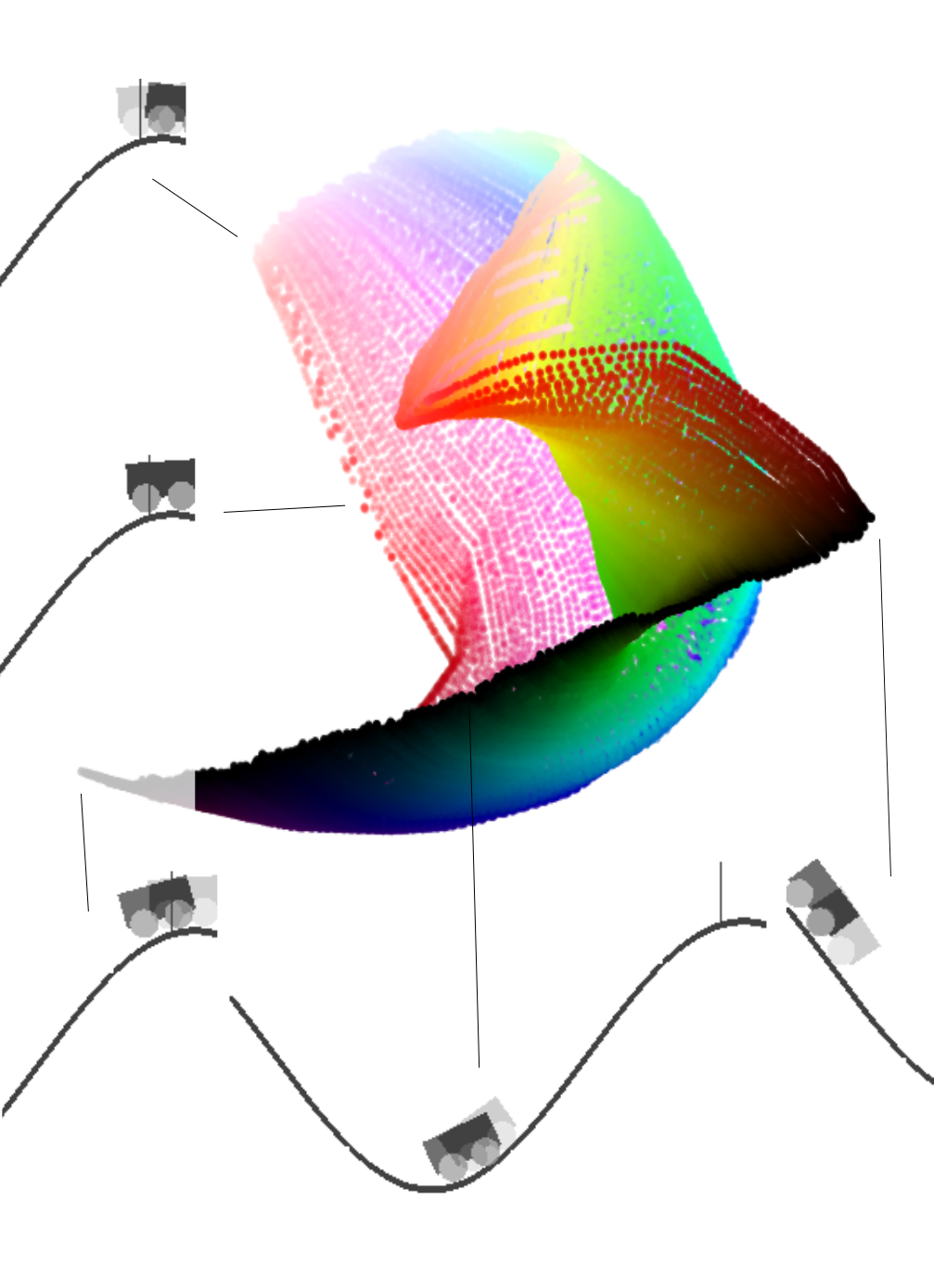}
  \end{minipage}\hfill
 \begin{minipage}[c]{0.65\linewidth}
  \caption{{
    The $E(3)$ equivariant representation of the mountain-car. Each state $x \in \Xs$ is a concatenation of two consecutive frames so as to inform about both position and velocity of the car. The colors encode the true position of the car, and the brightness of the colors shows the positive-negative velocity. The transformation $t_\Xs(\gg, x)$ changes the velocity through positive/negative acceleration.
  By preserving the distance between pairs of instances in which the same acceleration is applied, transformation coding is able to recover a manifold that is parameters by velocity and location.
  }}
  \label{fig:mountain-car}
  \end{minipage}
\end{figure*}

\section{Additional Experiment: the Mountain Car}
\label{appdx:MountainCar}
\begin{wrapfigure}{r}{0.15\linewidth}
\vspace{-5em}
  \begin{center}
    \includegraphics[width=\linewidth,trim={70 100 50 50},clip]{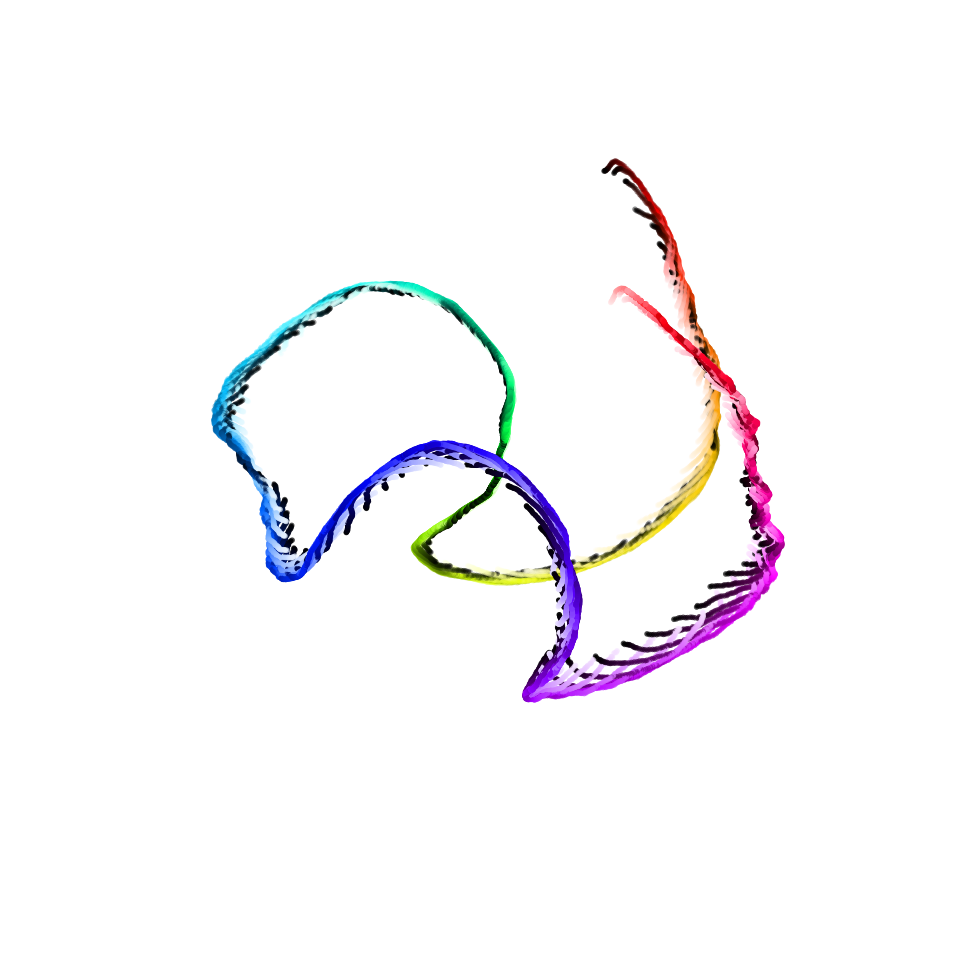}
  \end{center}
  \vspace*{-1em}
  \caption{VAE embedding for the Mountain Car example; compare againts \cref{fig:mountain-car}}\label{fig:vae-mountaincar}
\end{wrapfigure}
In the Mountain Car environment, an observation $x$ consists of two consecutive frames that have been grayscaled and downsampled to $32 \times 32$ pixels. The action-space of the environment is a range of accelerations that can be applied to the car. \cref{fig:mountain-car} shows the learned embedding in $\Reals^3$. \cref{fig:mountain-car} shows the learned embedding in $\Reals^3$.  Colors show the change in the location and the brightness shows the velocity. The learned representaion is quite intuitive, and the model learns to parameterize the manifold using the location and velocity of the car.

\section{Addition of Invariant Features} 
While we focused on equivariant codes, one may also consider an invariant component in the code which can account for variations in the data that are not due to transformations -- that is, we have $f: \Xs \to \Zs \times \Ys$, where $\Ys$ is the invariant part of the code that identifies distinct \emph{orbits}. Let $f^{\text{inv.}}: \Xs \to \Ys$ denote the invariant component of $f$. To learn $f^{\text{inv.}}$, one could use a loss of the form $\ell(f^{\text{inv.}}(x) - f^{\text{inv.}}(t_\Xs(\gg, x)))$ which enforces invariance for points on the same orbit. At the same time, an injection loss, similar to those of~\cref{sec:cheap}, pushes apart the points that are not in the same orbit. This invariant component is therefore very similar to what is used in contrastive coding.


\section{Effect of Transformations on the Embedding Manifold}
\begin{wrapfigure}{r}{0.4\linewidth}
\vspace*{-1em}
  \begin{center}
    \includegraphics[width=0.7\linewidth]{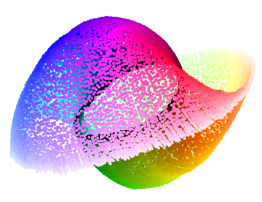}
  \end{center}
  \caption{{Alternative embedding for the pendulum.}}\label{fig:alternative}
  \vspace*{-2em}
\end{wrapfigure}

When using transformation coding, 
the transformation can have a complex relationship with the ideal parameterizations of the manifold.  For example, the location and velocity of the mountain-car or the pendulum (as ideal parameters for the manifold), are non-trivially related to the action that accelerates the movement.  
A natural question here is about the effect of the choice of transformation on the manifold. In practice, we observe that, in low-dimensional embedding, the geometry of the manifold is quite sensitive to the choice of transformation.
As an example,~\cref{fig:alternative} presents an alternative embedding for the pendulum of~\cref{fig:pendulum} obtained by simply decreasing the amount of time between taking an action, and observing its outcome from $\delta t=.05 \to \delta t = .01$. Whether or not this (potential) sensitivity is a bug or feature may depend on the application setting.

\section{Implementation Details}
\label{appdx:Implementation}
\textbf{Pendulum and Mountain Car}
We use the same setting for both of these environments. The neural network first applies three convolutional layers with $3 \times 3$ kernels, 24 output channels, and ``same'' padding, each followed by $2 \times 2$ max pooling and ReLU activation. Finally, a Multi-Layer Perceptron (MLP) with a single hidden layer of size 128 and ReLU activation is applied to embed the representation into $\Reals^3$. The model was trained for 5000 steps with the Adam optimizer set to a learning rate of $10^{-3}$. The log-barrier coefficient was set to 1, and we used a weight decay coefficient of $10^{-7}$.

\textbf{Bump World}
The neural network used for conformal coding experiments in \cref{exp:conformal_bumpworld} is a 4-layer MLP with hidden layers of size 128 and ReLU activation functions. The embedding space is $\Reals^4$ which is then randomly projected to $\Reals^3$ for visualization. The log-barrier coefficient and weight decay coefficient were both set to $10^{-7}$. The mini-batches used for training consist of 64 randomly sampled observations from the environment and their transformations via 15 randomly sampled transformations ($16 \times 64$ samples in total). The model was trained for $10,000$ steps with the Adam optimizer set to a learning rate of $10^{-3}$.

The neural network used in active and passive decomposition experiments in \cref{exp:decom_bump world} is a 3-layer MLP with hidden layers of size 128 and ELU activation functions. The embedding space is $\Reals^4$ which we interpret as two $\Reals^2$s. We optimize the model for 5000 steps using the Adam optimizer with an initial learning rate of $10^{-2}$ which is halved every 1000. We use a barrier coefficient of 1 and a weight decay coefficient of $10^{-5}$.
The mini-batches used for training consist of 64 randomly sampled observations from the environment and their transformations via 7 randomly sampled transformations ($8 \times 64$ samples in total).

\textbf{Gym Mini-world}
The neural network architecture and training settings are similar to those of the pendulum and mountain car experiments.

\textbf{Rotating chair}
 The neural network used for the Chair dataset first applies three convolutional layers with $3 \times 3$ kernels, increasing number of output channels from 16 to 32 to 64, 1 padding and stride of 2. It is followed by a ReLU activation. Finally, a Multi-Layer Perceptron (MLP) with a single hidden layer of size 128 and ReLU activation is applied to embed the representation into $\Reals^{16}$. The model was trained with 10000 unique rotations repeated with multiple initial points with the Adam optimizer set to a learning rate of $10^{-3}$. We stick to small rotation angles to make the setup similar to dynamical environments. The log-barrier loss coefficient was set to 1, and we used a weight decay coefficient of $10^{-7}$.

\textbf{Pong and Space Invaders}
The neural network used for the Chair dataset first applies three convolutional layers with $7 \times 7$ , $5 \times 5$ and $3 \times 3$ kernels. We again increase the number of output channels from 16 to 32 to 64 and use a padding of 1 and stride of 2 for all of them. The convolutions are followed by ReLU activations. Finally, a linear layer is applied to embed the representation into $\Reals^{32}$. We train the encoder with samples collected from around 100k environment steps. The log-barrier loss coefficient was again set to 1.

\textbf{Pendulum, Reacher and Swimmer}
The encoder and pre-training setting used for this part is similar to the one used for Pong and space invaders. We use an MLP to learn the policy from the extracted features. We use stable baselines 3 for our PPO implementation. \cite{stable-baselines3}.

\end{document}